\documentclass[12pt,journal,compsoc,onecolumn]{IEEEtran}
\usepackage{amsmath,epsfig}
 \usepackage{lingmacros}
 \usepackage{tree-dvips}
\usepackage{amsmath} 
\usepackage{amssymb}
\usepackage{pifont}
\usepackage[noadjust]{cite}
\usepackage{multirow}

\newlength{\bibitemsep}
\newlength{\bibparskip}
\let\oldthebibliography\thebibliography
\renewcommand\thebibliography[1]{
  \oldthebibliography{#1}
  \setlength{\parskip}{\bibitemsep}
  \setlength{\itemsep}{\bibparskip}
}

\def\x{{\mathbf x}}

\def\X{{\cal X}}

\def\Y{{\cal Y}}
\def\S{{\cal I}}
\def\C{{\cal C}}
\def\x{{\bf x}}

\def\D{{\cal D}}

\def\C{{\bf C}}
\def\F{{\bf  F}}
\def\DD{{\bf D}}
\def\tr{{\bf tr}}
\def\1{{\bf 1}}

\def\x{{\bf x}}  
\def\Y{{\cal  Y}}  

\def\A{{\cal A}}
\def\T{{T}}

\usepackage[ruled,vlined]{algorithm2e}

\title{Reinforcement-based Display-size Selection for Frugal  Satellite Image Change Detection} 
\author{Hichem Sahbi \\
$ $ \\
Sorbonne University, CNRS, LIP6,  F-75005, Paris, France 
 }

\begin{document}
 \maketitle
\begin{abstract}
We  introduce a novel interactive satellite image change detection algorithm based on active learning.  The proposed method is  iterative   and consists in frugally probing the user (oracle)  about  the labels of the most critical images,  and according to the oracle's annotations,  it updates change detection results.  First,  we   consider a probabilistic framework which  assigns to each unlabeled sample a relevance measure modeling how critical is that sample when training  change detection functions.   We obtain these relevance measures  by minimizing an objective function mixing diversity, representativity and uncertainty.  These criteria  when combined allow exploring different data modes and also refining change detections.  Then,  we further explore the potential of this objective function,  by considering  a reinforcement learning approach that  finds the best combination of diversity, representativity and uncertainty as well as display-sizes through  active learning iterations, leading to better generalization as shown through  experiments in interactive satellite image change detection. 
\end{abstract}

\section{Introduction}
\label{sec:intro}
 Satellite image change detection  aims at localizing {\it instances} of relevant (targeted) changes into a given scene captured at an instant $t_1$ w.r.t. the same scene taken at an earlier instant $t_0$ \cite{ref7,ref9,ref11,ref13}. These instances may correspond to {\it newly} appearing or disappearing entities due to infrastructure destruction after natural hazards (flash-floods, tornadoes, earthquakes, etc.) \cite{ref4,ref5}. This problem is highly challenging due to the eclectic properties of relevant changes and also the high variability of irrelevant  ones (illumination, occlusions, etc). In related state-of-the art, irrelevant changes are either removed by correcting and normalizing the radiometric effects in satellite images~\cite{ref14,ref15,ref17,ref20,refffabc8} or by considering these changes as a part of appearance modeling~\cite{ref21,ref25,ref26,ref27,ref28,refffabc7,refffabc5,refffabc6}. This modeling seeks to design statistical, machine and deep learning algorithms~\cite{refffabc3,refffabc4} that discriminate between relevant and irrelevant changes. However, the success of these algorithms is highly reliant on the availability of enough labeled data that capture all the inherent variability of relevant and irrelevant changes. In practice, labeled data are scarce and, even when available, the relevance of changes is subjective (user-dependent) and this renders the task of change detection very challenging. \\
 
\indent Machine learning methods that attenuate the effect of label scarcity are usually based on zero/few shot, self-supervised and active learning~\cite{reff45,refff2,reff2,reff16,reff15,reff53,reff12,reff74,reff58,reff1,refffabc0,refffabc1}. In particular, active learning is highly effective and makes it possible to model the user's intention --- about targeted changes --- interactively by (i) showing the most critical unlabeled data (dubbed as displays) to the user/oracle, (ii) probing the latter about their relevance, and (iii) updating change detection results~\cite{refff33333}. Finding the most suitable display to {\it frugally} probe the oracle is usually achieved by maximizing the diversity, the representativity and the uncertainty  of the selected unlabeled data \cite{reff13}. Nonetheless, knowing a priori which sequence of display strategies (diversity, representativity and uncertainty) to apply, and how many samples to label while trading off  accuracy and training/label efficiency, through all the iterations of active learning, is highly combinatorial. Furthermore, under the frugal labeling regime,  labeled validation data are scarce to make the design of optimal display strategies statistically meaningful. \\

\indent In this paper, we investigate  the design of ``optimal'' strategies for interactive satellite image change detection using reinforcement learning (RL). First, we consider a  probabilistic framework that assigns for each unlabeled sample a relevance degree which measures how important is that sample when learning change detection criteria. These relevances are obtained by solving a constrained objective function that mixes diversity, representativity and ambiguity. Then, we design different reward functions that allow selecting and combining the best sequence of actions (diversity,  representativity and ambiguity as well as display sizes) which efficiently tackles the combinatorial aspect of theses actions and ultimately leads to optimal change detection performances. In particular, display-size selection is critical in order to trade-off display model effectiveness and efficiency, i.e.,  fine-grained (more effective) vs coarse-grained (more efficient) display model update.  Finally, we show extensive experiments on the challenging task of interactive satellite image change detection and the outperformance of our proposed RL-based design against the related work.

\section{Proposed method}
\label{sec:proposed}

Considering $\X=\{\x_i=(p_i,q_i)\}_{i=1}^n$ as a set of aligned patch pairs taken from two registered satellite images ${\cal I}_0=\{p_1,\dots,p_n\}$, ${\cal I}_1=\{q_1,\dots,q_n\}$ captured at two different instants. Initially, the labels of $\X$ are unknown, and our goal is train a change decision function $g(.)$ by allowing the oracle\footnote{The oracle is an expert that labels a subset of images as changes / no-changes.} to interactively label a very small fraction of  $\X$ --- as a sequence of subsets $\{\D_t\}_{t\in \mathbb{N}^+}\subseteq \X$ (dubbed as displays) --- and by training $g(.)$ on $\{(\D_t,\Y_t)\}_t$, being $\Y_t$ the labels assigned by the oracle to $\D_t$. This process is known as active learning.  At $t=0$, the initial display $\D_0$ is  uniformly sampled at random, and used to build change detection criteria by alternately (i) training $g_t(.)$ using $\bigcup_{\tau=1}^t (\D_\tau,\Y_\tau)$ where the subscript in  $g_t(.)$ refers to the decision function at  iteration $t$, and (ii) selecting the following display $\D_{t+1} \subset {\X}-\bigcup_{\tau=1}^t \D_\tau$ that possibly increases the generalization performances of the subsequent classifier $g_{t+1}(.)$. These steps (i)+(ii) are iteratively applied till exhausting a {\it fixed} labeling budget. It is clear that one cannot combinatorially select all the possible displays  $\D_{t+1} \subset {\X}-\bigcup_{\tau=1}^t \D_\tau$, train classifiers\footnote{In this paper, graph convolutional networks are used for classification.} on top of these subsets and keep the best display (i.e., the one that maximizes classification performances); on the one hand, labeling should be frugally achieved, and on the other hand, testing all the possible subsets $\D_{t+1}$ cannot be exhaustively achieved. Hence, relevant display selection strategies --- related to active learning --- should instead be considered~\cite{reff1}. In what follows, we introduce our novel display selection strategies which allow finding the most representative, diverse and ambiguous data --- together with the optimal display sizes ---  that eventually lead to the most discriminating change detection criteria while being label-efficient, and this constitutes the main contribution of this paper.

\subsection{Display Selection}
We assign for each unlabeled sample $\x_i$  a relevance degree $\mu_i$ that measures how important is $\x_i$ in {\it shaping} the next display $\D_{t+1}$. Equivalently, $\D_{t+1}$ will correspond to the unlabeled data in $\{\x_i\}_i \subset \X$ with the highest $\{\mu_i\}_i$. Let $\mu \in \mathbb{R}^{|{\X}|}$ be a vector of these relevances  $\{\mu_i\}_i$, the vector $\mu$ is found by optimizing the following criteria \\

\noindent {\bf -Representativity.} Let $\{h_k\}_k$ be a partition of data in $\X$ into K-clusters obtained with K-means. We measure representativity of our selected subset in $\X$ using  $\sum_i \sum_k  1_{\{\x_i \in h_k \}} \mu_i d_{ik}^2$; being $d_{ik}^2$  the euclidean distance between $\x_i$ and $k^{\textrm{th}}$ cluster centroid of $h_k$. This criterion captures how close is each $\x_i$ w.r.t. the centroid of its cluster, so this term reaches its smallest value when all the selected samples coincide with their centroids.  \\ 

\noindent {\bf -Diversity.} We measure the diversity of the selected samples using $\sum_{k}   [\sum_i 1_{\{\x_i \in h_k \}} \mu_i] \log [\sum_i 1_{\{\x_i \in h_k \}} \mu_i]$ as the entropy of the probability distribution of the cluster partition $\{h_k\}_k$; this measure is minimized when the selected samples belong to different clusters and vice-versa.  \\ 

\noindent {\bf -Ambiguity.} Let  $\hat{g}_t \in [0,1]$ be a normalized version\footnote{This normalization is obtained, in practice, using softmax.} of $g_t$, we measure the ambiguity of the selected samples using  $\sum_i \mu_i [\hat{g}_t(\x_i) \log \hat{g}_t(\x_i)  + (1-\hat{g}_t(\x_i)) \log (1-\hat{g}_t(\x_i))]$ which corresponds to the entropy of the scoring function on the selected data in $\D_{t+1}$. This term reaches its smallest value when data are evenly scored w.r.t. the ``change'' and ``no-change'' classes.\\ 

\noindent {\bf -Combination.} Considering the matrix forms of the three aforementioned criteria, one may combine them as
  \def\diag{{\textrm{diag}}}
 {\begin{equation}\label{eq0}
  \begin{array}{l}
    \displaystyle    \min_{\mu \geq 0, \|\mu\|_1=1}  \eta \ \tr\big(\diag(\mu' [\C \circ \DD])\big) + \alpha  \ [\C' \mu]' \log [\C' \mu] \\
             \ \ \ \ \ \ \  \ \ \ \ \ \ \ \ \ \ \    \ \   + \beta \  \tr\big(\diag(\mu' [\F \circ \log \F]) \big) + \gamma \mu' \log \mu, 
 \end{array}  
\end{equation}}

\noindent here $\eta$, $\alpha$, $\beta$, $\gamma \geq 0$,  and $\circ$, $'$ are respectively the Hadamard product and the matrix transpose, $\|.\|_1$ is the $\ell_1$ norm, $\log$ is applied entry-wise, and $\diag$ maps a vector to a diagonal matrix.  In the above objective function (i) $\DD \in \mathbb{R}^{|\X| \times K}$ and $\DD_{ik}=d_{ik}^2$ is the euclidean distance between $\x_i$ and the $k^{\textrm{th}}$ cluster centroid, (ii) $\C \in \mathbb{R}^{|\X| \times K}$ is the indicator matrix with each entry  $\C_{ik}=1$ iff $\x_i$ belongs to the $k^{\textrm{th}}$ cluster ($0$ otherwise),  and (iii) $\F \in \mathbb{R}^{|\X| \times 2}$ is a scoring matrix with $(\F_{i1},\F_{i2})=(\hat{g}_t(\x_i),1-\hat{g}_t(\x_i))$. The fourth term is added and it is related to the {\it cardinality} of $\D_{t+1}$, measured by the entropy of the distribution $\mu$ (this term also acts as a regularizer), and equality/inequality constraints guarantee that $\mu$ form a probability distribution.   Considering $\1_{nc}$, $\1_{K}$ as vectors of $nc$ and $K$ ones respectively (with $nc=2$ in practice),  one may show that the solution of Eq.~\ref{eq0} is given as $\mu^{(\tau+1)} :=\displaystyle \hat{\mu}^{(\tau+1)}/\|\hat{\mu}^{(\tau+1)}\|_1$, with  $\hat{\mu}^{(\tau+1)}$ being
\begin{equation}\label{eq2}
 \exp\bigg\{-\frac{\eta}{\gamma} (\DD\circ \C)\1_K\bigg\} \ \circ \ \exp\bigg\{-\frac{\alpha}{\gamma} \C (\log[\C' {\mu}^{(\tau)}]+\1_K)\bigg\} \ \circ \  \exp\bigg\{-\frac{\beta}{\gamma} (\F \ \circ \  \log \F)\1_{nc}\bigg\}. 
\end{equation}
 
 \noindent As shown subsequently and  later in experiments, the setting of the mixing hyper-parameters ($\eta$, $\alpha$, $\beta$, $\gamma$) is crucial for the success of display selection.  For instance, putting emphasis on diversity (i.e.,  $\alpha>0$) results into exploration of class modes while a focus on ambiguity (i.e.,  $\beta>0$) locally refines the trained decision functions.  A suitable balance between exploration and local refinement of the learned decision functions should be achieved by selecting the best combination of these hyper-parameters through active learning iterations.  Nevertheless, since labeling is sparingly achieved by the oracle,  no sufficiently large validation sets could be made  available beforehand to accurately set these hyper-parameters.\\
 
\noindent Note that optimizing $\gamma$ allows defining the size of $\D_{t+1}$. Small $\gamma$-values make the distribution in Eq.~\ref{eq2} peaked and only a few samples $\{\x_i\}_i$ will have nonzero values $\{\mu_i\}_i$. Therefore, only a few data will be selected and labeled by the oracle resulting into multiple (expensive) re-running of the active learning iterations in order to reach the targeted labeling budget, but this is achieved at the benefit of fine-grained and hence more accurate updates of the learned display model. In contrast, large $\gamma$-values make the distribution flat and hence the display $\D_{t+1}$ larger; thus, data will be labeled {\it at whole}   by the oracle and this allows {\it reaching the targeted labeling budget and training the underlying classifiers faster}, but at the detriment of coarse-grained (less accurate) updates of the learned display model,  ultimately leading to less accurate subsequent classifiers.  Finding the best $\gamma$ that eventually corresponds to a predefined labeling budget, while guaranteeing a sufficiently fine-grained update of the display model, is challenging and also less intuitive when relying on $\gamma$. In what follows, we omit $\gamma$ in Eq.~\ref{eq2} by fixing its value to $1$, and we consider an alternative solution for display-size selection based on reinforcement learning. 
\subsection{RL-based Display Selection}\label{rlbased}
\indent Let $\Lambda_\alpha$,  $\Lambda_\beta$, $\Lambda_\eta$ denote the parameter spaces associated to $\alpha$,  $\beta$, $ \eta$ respectively, let $\Lambda_s$ be the parameter space associated to the display size,  and let $\Lambda$  be the underlying   Cartesian product.  For any $\lambda_{t+1}=(\alpha_{t+1},\beta_{t+1},\eta_{t+1},|\D_{t+1}|) \in \Lambda$ ($\lambda_{t+1}$ written for short as $\lambda$),  one may obtain a display (now rewritten as  $\D^\lambda_{t+1}$) by solving   Eq.~\ref{eq0}.  In order to find  the best configuration $\lambda^*$ that yields an ``optimal'' display,  we model hyper-parameter selection as a Markov Decision Process (MDP). An MDP based RL corresponds to a tuple $\langle \S,\A,R,\T,\delta\rangle $ with $\S$ being a state set, $\A$ an action set,  $R: \S \times \A \mapsto \mathbb{R}$ an immediate  reward function,  $T: \S\times \A \mapsto \S$ a transition function and $\delta$ a discount factor \cite{refff333334}.   RL consists in running a sequence of actions from $\A$ with the goal of maximizing an expected discounted reward by following a stochastic policy, $\pi: \S \mapsto \A$;  this leads to the true state-action value as  $Q(s,a) = E_\pi \left[  \sum_{k=0}^\infty  \delta^k {r}_k | S_0,=s,  A_0=a \right]$,
here  $E_\pi$ denotes the expectation w.r.t.  $\pi$,   ${r}_k$ is the immediate reward at the $k^{\textrm{th}}$ step of RL,   $S_0$ an initial state,  $A_0$ an initial action  and $\delta \in [0,1]$ is a discount factor that balances between immediate and future rewards. The goal of the optimal policy is to select actions that maximize the  discounted cumulative reward; i.e., $\pi_{*}(s) \leftarrow \arg\max_{a} Q(s,a)$.  One of the most used methods to solve this type of RL problems is Q-learning \cite{refff333335}, which directly estimates the optimal value function and obeys the fundamental identity, the Bellman equation 
\begin{equation}
Q_*(s,a) = E_\pi \left[ {R}(s,a) + \delta \max_{a'} Q_*(s',a') | S_0=s,A_0=a \right],
\end{equation} 
with $s'=T(s,a)$ and ${ R}(s,a)$ is again the immediate reward.    We consider in our hyper-parameter optimization, a stateless version,  so $Q(s,a)$ and $R(s,a)$ are rewritten simply as $Q(a)$, $R(a)$ respectively. In this configuration,  the  parameter space $\Lambda$ is equal to $\{0,1\}^3\backslash(0,0,0) \times \{-1,0,+1\}$  so the underlying action set $\A$ corresponds to 7 possible binary (zero / non-zero) settings of  $\alpha, \beta,\eta$ {\it combined} with 3 possible updates of display sizes: here $-1$, $0$, and $+1$ respectively stand for display-size decrease,  freeze and increase.  In total  $7 \times 3$ possible actions are used in order to update the displays with Eq.~\ref{eq0}. We consider an adversarial immediate reward function $R_a$ that scores a given action  (and hence the underlying configuration $\lambda \in \Lambda$)   proportionally to the error rates of $g_t(\D_{t+1}^{\lambda})$; put differently, the display $\D_{t+1}^{\lambda}$ is  selected in order to challenge (the most) the current classifier $g_t$, leading to a better estimate of  $g_{t+1}$. We combine $R_a$ with  an efficiency reward $R_s$ which is proportional to the size of the display. With $R_s$, larger displays are encouraged (and hence few active learning iterations) but this may lead to coarse-grained updates of the display model and eventually to low accuracy. Hence, with $R_s$ and $R_a$, our RL-based design balances efficiency (by selecting the largest displays that lead to few active learning iterations) and accuracy (by achieving fine-grained display model updates)  as shown subsequently in experiments.  
 
 \section{Experiments}\label{exp}
\indent {\bf Eval dataset.}  We evaluate the accuracy of our RL-based interactive change detection algorithm using the Jefferson dataset.  The latter consists of $2,200$ non-overlapping ($30\times30$ RGB) patch pairs taken from (bi-temporal) GeoEye-1 satellite images of $2, 400 \times  1, 652$ pixels with a spatial resolution of 1.65m/pixel.  These patch pairs pave a large area from Jefferson (Alabama) in 2010 and in 2011.  These images show several damages caused by tornadoes (building destruction, debris on roads, etc) as well as no-changes including irrelevant ones (clouds, etc). In this dataset $2,161$ patch pairs correspond to negative data and only $39$ pairs to positive, so $<2\%$ of these data correspond to relevant changes and this makes their detection very challenging. In our experiments, half of the patch pairs are used for training and the remaining ones for testing.  We measure the accuracy of change detection using the equal error rate (EER); the latter is a balanced generalization error that evenly weights errors in  the positive and negative classes. Smaller EERs imply better performances.  \\ 
 
   \begin{table}[h]
   \begin{center}
 \resizebox{0.80\columnwidth}{!}{
 \begin{tabular}{c|c||cccccccc}
\multirow{10}{*}{\rotatebox{90}{Display size = 8}}  & Iter  &2 & 4& 6& 8& 10  & 12 &14 & 16 \\
&   Samp\%  & 1.45&   2.90&   4.36&   5.81&    7.27&    8.72&   10.18& \bf11.63  \\
\cline{2-10}
 &  rep     & 42.48  &  13. 21 &   10.98  & 09.75  &  09.53  &  08.72  &  08.35  & 07.55   \\                      
 & amb    & 45.87 & 43.92  &  43.63  & 41.13  & 39.36  &  38.05 & 34.67  &  32.04   \\                       
  & div     & 42.39  & 34.37  & 26.22   & 16.55   & 11.62  &  08.73  & 05.74  & 04.36  \\                     
    & amb+rep      & 41.68  & 21.97  &  14.70  &  12.98  &  11.42  &  11.32  &  06.37 &  05.88   \\                      
    &  div+amb   &41.06  &  35.56  & 23.16  & 17.76  & 13.26  &  10.08  & 06.51  &05.62  \\                       
      &  div+rep   & 41.11  &  25.13  & 13.94  &  09.98  &  09.43  &  07.53  &  04.94  &  04.50  \\                       
      &  all (flat) &  42.17  &  27.52 &  12.94  &   09.35   &07.78  &  06.87  &  04.82  &  04.27  \\
       &   RL-based  & 47.88 &   10.88  & 03.59  & 02.76 &   02.07 &   01.98  &  01.89 &   \bf02.21 \\
  \hline
 \hline
  \multirow{10}{*}{\rotatebox{90}{Display size = 16}} & Iter &2 & 3& 4& 5& 6& 7& 8& --- \\
 &  Samp\%  &2.90 & 4.36& 5.81& 7.27& 8.72& 10.18& \bf11.63&  ---\\
 \cline{2-10}
 & rep     &  23.48   & 12.12  &  09.79  &  08.40   & 08.89  &  08.11 &   07.47 &   ---    \\
   & amb     & 45.57   & 44.22 &  42.88  &  42.52  &  41.33  &  39.96 &   38.64  &  ---  \\
    &div     & 33.60  &  23.62 &   21.19  &  18.93 &   16.71  &  07.43  &  04.67  & --- \\
 & amb+rep      &  34.64   & 18.93  &  10.65   & 13.52   & 10.49   & 10.23   & 07.94   & ---  \\
  &  div+amb    &  34.37 &   24.76  &  18.80  &  13.54  &  15.14 &   10.24  &  08.75 &  --- \\
  & div+rep      & 33.34  &  18.83  &  11.33 &   12.42  &  10.58  &  09.59   & 09.07  & ---  \\
  & all (flat)   & 33.16   & 19.95  &  11.47   & 11.30  &  09.84   & 07.75  &  07.12  &  ---  \\
   &   RL-based   &  13.59   & 08.20  &  05.16  &  05.44  &  05.12 &   03.64  &  \bf04.56 &  --- \\    
        \hline
 \hline            
 \multirow{10}{*}{\rotatebox{90}{Display size = 32}}    & Iter  &2 & 3& 4& ---& ---&---& ---& --  \\
      &  Samp\%  &           5.81 &    8.72 &   \bf11.63 &   ---&    --- & ---& --- & ---   \\    
      \cline{2-10}
  &rep  &            11.73  & 09.86  &  10.37  &  ---         &    --- & ---& --- & --- \\                   
  & amb  &         44.74  &  43.40   & 41.85  &  ---       &    --- & ---& --- & ---  \\                   
   & div  &           22.39   & 15.12   & 11.56   & ---        &    --- & ---& --- & ---    \\                   
     & amb+rep  &     18.73 &   09.74 &  08.11  &  ---   &    --- & ---& --- & --- \\                   
     & div+amb  &     20.26  &  11.64  &  07.77  &  --- &    --- & ---& --- & --- \\                   
    & div+rep  &       18.71  &  10.70  &  08.28   & --- &    --- & ---& --- & --- \\                   
    & all (flat)  &         20.91  &  09.65  &  07.10  &  ---&    --- & ---& --- & --- \\                                  
  & RL-based &      16.82  &  08.02   &  \bf06.73   & ---&         --- & ---& --- & ---  \\
          \hline
          \hline            
      \multirow{3}{*}{\rotatebox{90}{{\scriptsize Adaptive}}} & Iter  &2 & 3& 4& 5& 6&7 & ---  &---  \\  
   &     Samp\%  &     3.18   &    4.90   &   6.54        &   8.09  &    9.72        & \bf11.45 &  --- & ---  \\
   \cline{2-10}
     & RL-based  & 43.04  &  37.05 &  02.07     &   01.47  & 01.43      &  \bf01.43  &  --- &---   \\
\end{tabular}}
 \caption{This table shows an ablation study of our display model. Here rep, amb and div stand for representativity, ambiguity and diversity respectively. These results are shown for different iterations $t$ (Iter) and the underlying sampling rates (Samp)  defined as $(\sum_{k=0}^{t-1} |\D_k|/(|\X|/2))\times 100$.  ``---'' stands for not applicable as the max number of sampled,  and labeled data by the oracle,  is reached.}\label{tab1} 
 \end{center}
 \end{table} 
\noindent {\bf Ablation study and impact of RL.} We first show an ablation study of our display selection model and mainly  the impact of ambiguity, representativity and diversity  criteria when taken individually and combined (for different display sizes).  From these results, we observe the positive impact of diversity at the early iterations of active learning, while the impact of ambiguity comes later in order to further refine the learned change detection functions. However, none of the settings (rows) in table~\ref{tab1}  obtains the best performance through all the iterations of active learning. For all display sizes, and for equivalent percentages of labeled data,  a better setting of the $\alpha$, $\beta$ and $\eta$ should be cycle-dependent using reinforcement learning (as described in section~\ref{rlbased}), and as also corroborated through performances shown in table~\ref{tab1}.  We also observe that fine-grained display model updates (i.e.,  with small display sizes) lead to better performances for equivalent labeled data; nonetheless, this is obtained at the expense of longer active learning iterations.  Moreover,  better performances are obtained with adaptive display updates using RL.   Indeed, it turns out that this adaptive RL-based  display update outperforms  the other combinations (including ``all'',  also referred to as ``flat'' as well as other RL-based settings with fixed display sizes), especially at the late iterations of change detection; we also observe that the number of iterations is also reasonably small.\\ 

\begin{figure}[tbp]
\center
\includegraphics[width=0.6\linewidth]{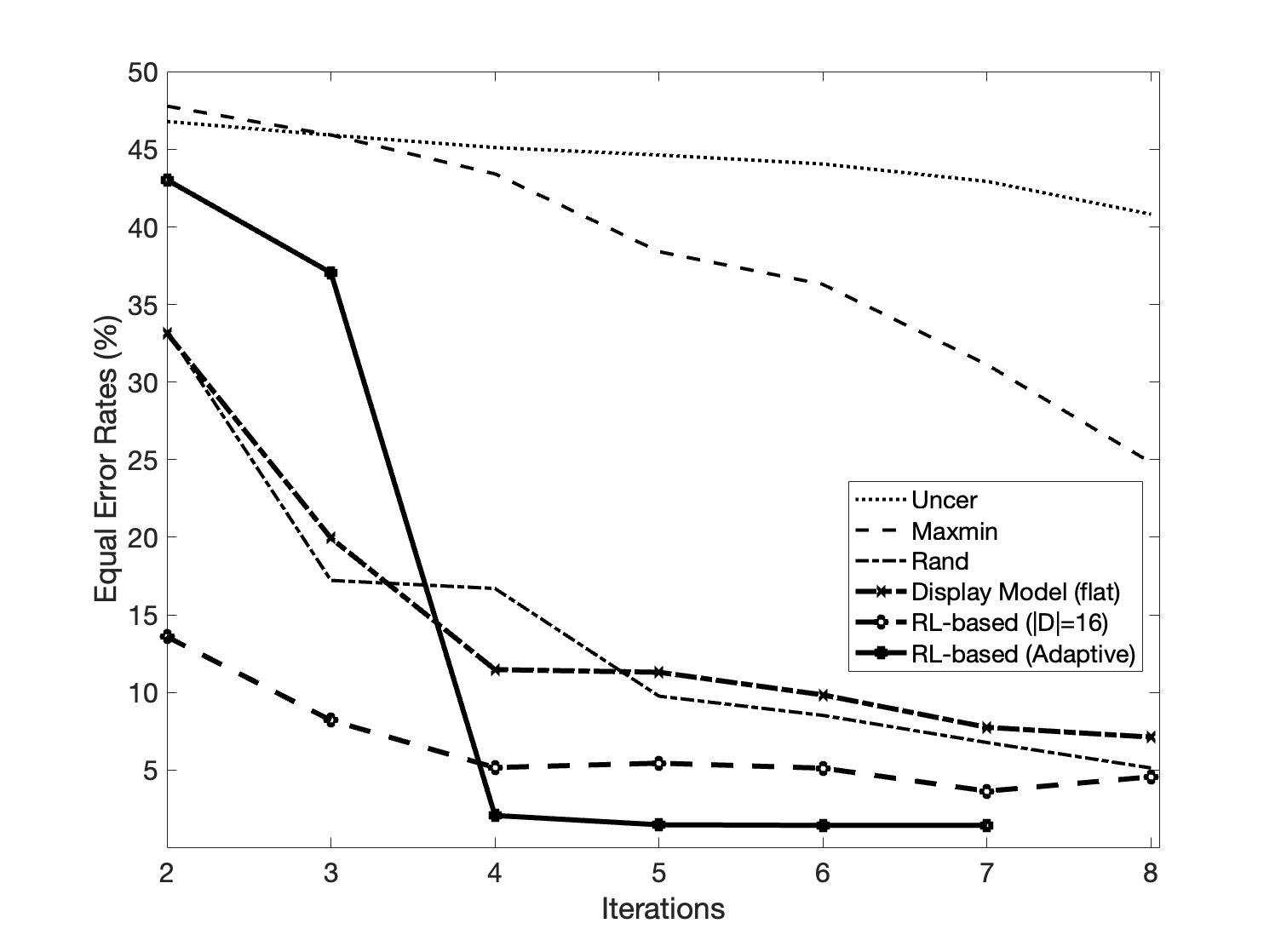}
 \caption{This figure shows a comparison of different sampling strategies w.r.t. different iterations (Iter) and the underlying sampling rates in table~\ref{tab1} (Samp). Here Uncer and Rand stand for uncertainty and random sampling respectively. Note that fully-supervised learning achieves an EER of $0.94 \%$. See again section~\ref{exp} for more details.}\label{tab2}\end{figure}

\noindent {\bf Extra comparison.} Figure.~\ref{tab2} shows extra comparisons of our RL-based display model w.r.t. different related display sampling techniques including {\it random, MaxMin and uncertainty}.  Random picks data from the unlabeled set whereas MaxMin greedily selects a sample $\x_i$  in ${\cal D}_{t+1}$ from the pool ${\X}\backslash \cup_{k=0}^t {\cal D}_k$ by maximizing its minimum distance w.r.t  $\cup_{k=0}^t {\cal D}_k$. We also compare our method w.r.t. uncertainty which consists in selecting samples in the display whose scores are the closest to zero (i.e., the most ambiguous). Finally, we also consider the fully supervised setting as an upper bound on performances; this configuration relies on the whole annotated training set and builds the learning model in one shot. \\ 
The EERs  in figure~\ref{tab2} show the positive impact of the proposed RL-based display model  (both with fixed and  adaptive update of the display size) against the related sampling strategies for different amounts of annotated data. The comparative methods are effective either at the early iterations of active learning (such as MaxMin and random which capture the diversity of data without being able to refine decision functions) or at the latest iterations (such as uncertainty which locally refines change detection functions but suffers from the lack of diversity). In contrast, our proposed RL-based design (particularly with adaptive display size selection) adapts the choice of these criteria as active learning cycles evolve, and thereby allows our interactive change detection to reach lower EERs and to overtake all the other strategies at the end of the iterative process.

\section{Conclusion} 
We introduce in this paper  a satellite image change detection algorithm based on active and reinforcement learning.  The strength of the proposed method resides in its ability to find and adapt --- both  display selection criteria and sizes ---  to the active learning iterations,  thereby leading to more informative subsequent displays and more accurate decision functions.  Extensive experiments conducted on the challenging task of change detection shows the accuracy and the out-performance of the proposed interactive method w.r.t. the related work.


\begin{thebibliography}{1}
\bibitem{aaaaa9}  H. Sahbi.  A particular Gaussian mixture model for clustering and its application to image retrieval.   Soft Computing 12 (7), 667-676
\end{thebibliography}
\end{document}